%% file: root.tex
\documentclass[letterpaper, 10 pt, conference]{ieeeconf-modified}
\IEEEoverridecommandlockouts                              

\usepackage{graphics} 
\usepackage{times} 
\usepackage{amsmath} 
\usepackage{amssymb}  

\usepackage{mathtools}
\usepackage{soul}  
\usepackage[table,xcdraw]{xcolor}

\usepackage[numbers]{natbib}
\usepackage{graphicx}
\usepackage{textcomp}
\usepackage{array}
\usepackage{multirow}
\usepackage{hyperref}
\usepackage[capitalize]{cleveref}

\usepackage{lipsum}  

\crefname{equation}{}{}  

\newcommand{\joints}{q}

\newcommand{\rel}[3]{{}^{#2}{#1}_{#3}}

\newcommand{\Trobot}{F}
\newcommand{\T}[2]{{}^{#1}T_{#2}}
\newcommand{\Tcamera}{\T{0}{\mathrm{C}}}
\newcommand{\TcameraInv}{\T{\mathrm{C}}{0}}

\newcommand{\funForward}{f}
\newcommand{\funMeasurement}{h}

\newcommand{\funDHpar}{\rho}
\newcommand{\funCameraProjection}{U}
\newcommand{\funCameraDistortion}{D}
\newcommand{\funCameraThreeTotTwo}{P}

\newcommand{\CALpar}{\Theta}
\newcommand{\CALparForward}{\CALpar_{\mathrm{f}}}
\newcommand{\CALparProjection}{\CALpar_{\mathrm{C}}}

\newcommand{\DHpar}{\rho}
\newcommand{\DHparR}{\rho_0}

\newcommand{\CPpar}{\kappa} 

\newcommand{\Masspar}{\nu}
\newcommand{\torque}{\tau}

\newcommand{\CameraCenterPoint}{c_{\mathrm{C}}}
\newcommand{\CameraFocalLength}{f_{\mathrm{C}}}
\newcommand{\CameraDistortion}{\xi_{\mathrm{C}}}
\newcommand{\markerProjection}{u}
\newcommand{\markerPoint}{x}

\newcommand{\DataSet}{S}

\newcommand{\gitlink}{https://dlr-alr.github.io/2022-humanoids-calibration}

\usepackage{fancyhdr}
\fancypagestyle{FirstPage}{
\lfoot{\fontsize{7}{7}\selectfont \begin{center}\copyright2022 IEEE. Personal use of this material is permitted. \\
Permission from IEEE must be obtained for all other uses, in any current or future media, including reprinting/republishing this material for advertising or promotional purposes, creating new collective works, for resale or redistribution to servers or lists, or reuse of any copyrighted component of this work in other works.\end{center}}
}

\title{\LARGE \bf
Self-Contained Calibration of an Elastic Humanoid Upper Body \\ 
Using Only a  Head-Mounted RGB Camera
}
\author{Johannes Tenhumberg$^{1,2}$\;\; Dominik Winkelbauer$^{1}$ \;\; Darius Burschka$^{3}$ \;\; Berthold Bäuml$^{1,2}$
\thanks{$^{1}$DLR Institute of Robotics \& Mechatronics, Germany;
$^{2}$Deggendorf Institute of Technology, Germany; 
$^{3}$Technical University of Munich, Germany}
\thanks{Contact: \tt\footnotesize johannes.tenhumberg@dlr.de}
\thanks{This work was partly funded by the Bavarian Ministry of Economic Affairs, Regional Development and Energy, within the projects SMiLE (LABAY97) and SMiLE2gether (LABAY102).}
}

\begin{document}
\maketitle
\thispagestyle{empty}
\pagestyle{empty}

\begin{abstract}
When a humanoid robot performs a manipulation task, it first makes a model of the world using its visual sensors and then plans the motion of its body in this model.
For this, precise calibration of the camera parameters and the kinematic tree is needed.
Besides the accuracy of the calibrated model, the calibration process should be fast and self-contained, i.e., no external measurement equipment should be used.
Therefore, we extend our prior work on calibrating the elastic upper body of DLR's Agile Justin by now using only its internal head-mounted RGB camera. 
We use simple visual markers at the ends of the kinematic chain and one in front of the robot, mounted on a pole, to get measurements for the whole kinematic tree.
To ensure that the task-relevant cartesian error at the end-effectors is minimized, we introduce virtual noise to fit our imperfect robot model so that the pixel error has a higher weight if the marker is further away from the camera.
This correction reduces the cartesian error by more than 20\,\%, resulting in a final accuracy of 3.9\,mm on average and 9.1\,mm in the worst case. 
This way, we achieve the same precision as in our previous work~\cite{Tenhumberg2021}, where an external cartesian tracking system was used.
\end{abstract}

\section{Introduction}
\thispagestyle{FirstPage}
When the humanoid robot Agile Justin performs a task, it first uses its internal camera to make a model of the world, including the poses of objects.
It then plans how to move its body in this world model to reach an object without obstacle collison or self-collisions.
The success of this look-and-move approach depends highly on the calibration of its cameras and the whole kinematic tree.
In the case of Agile Justin, the deviation from the nominal geometric kinematics is as large as 61\,mm.
This significant error makes it necessary to add safety margins for the collisions, and robust and precise manipulation is almost impossible.

In our previous paper~\cite{Tenhumberg2021}, we derived a model with elasticities for the humanoid and showed how to use it efficiently inside an optimization-based planner.
The calibration of the model was based on the cartesian measurements of an external tracking system where tracking targets were mounted on the two end effectors.
Using an external tracking system poses two main problems. 
First, calibration is only possible when in the lab. 
Second, the internal camera is not incorporated in the calibration, although it is used when performing manipulation tasks, limiting the accuracy.

In this paper, we show that the elastic model of a humanoid robot can be calibrated using only its head-mounted RGB camera and simple markers on the two end effectors and one in front of the robot.
The main contributions are:

\begin{figure}[tb]
	\centering
	\includegraphics[width=0.6\linewidth]{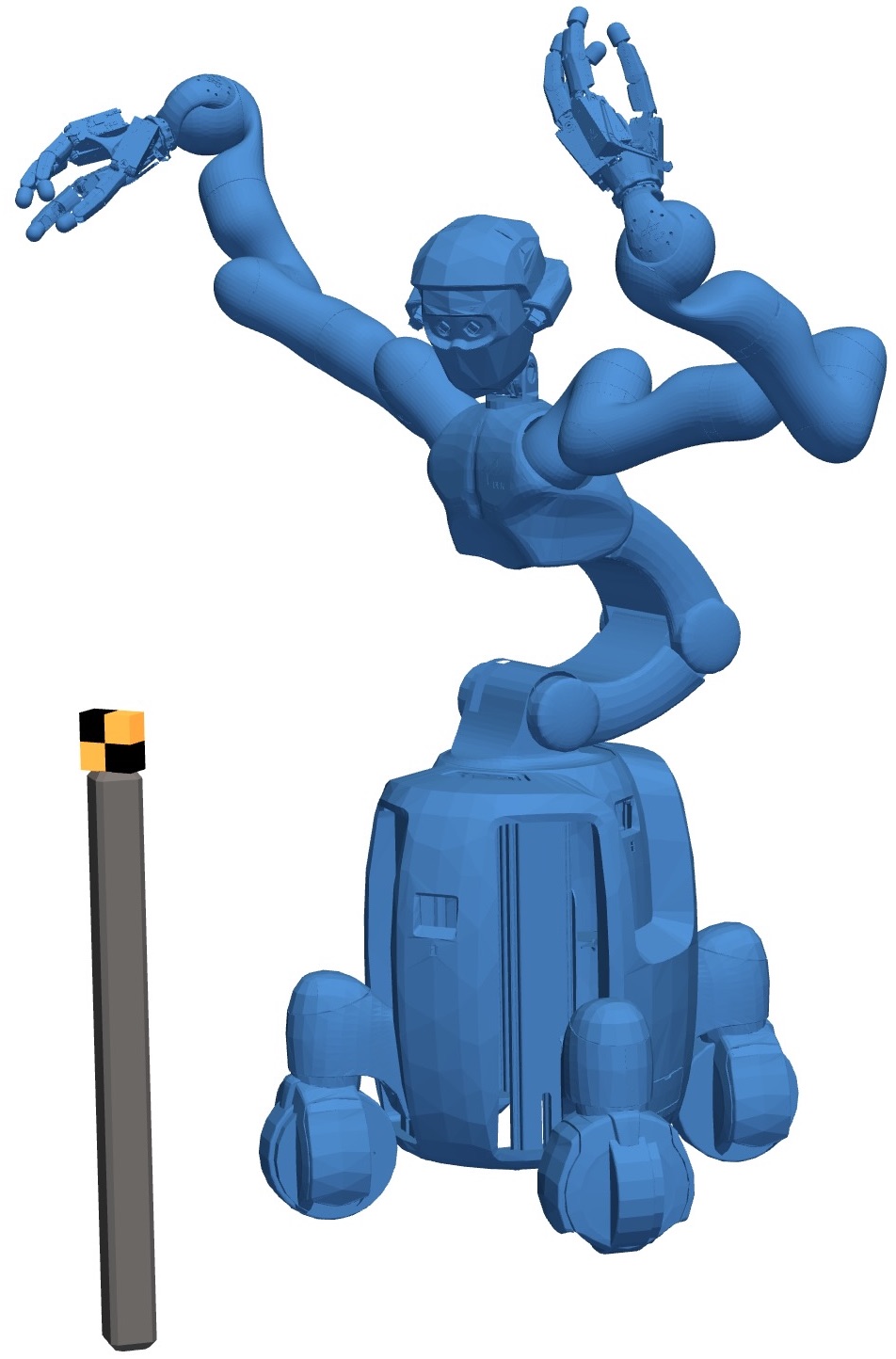}	
	\caption{ DLRs's Agile Justin~\cite{Bauml2014} collecting measurements using its head-mounted RGB camera for the calibration of its elastic forward kinematics as well as the camera's intrinsic and extrinsic parameters. Only simple markers on both hands as well as the depicted marker mounted on a pole are used. As described in \cref{sec:Configuration-Selection}, we select filtered random configurations to identify the robot in its whole work space.}
	\label{fig:JustinPoleDance}
\end{figure}

\begin{itemize}
\item We perform a self-contained robot calibration using only the head-mounted $640\!\times\!480$ RGB camera, i.e., without any external measurement equipment.
\item The complete kinematic tree, including the torso, left and right arms, the neck, and the camera, are calibrated.
The model has 129 free parameters, including the DH parameters, joint and lateral elasticities, and the extrinsic and intrinsic parameters of the camera.
\item We show that directly minimizing the error between measured and reprojected pixel coordinates of the markers results in non-optimal cartesian precision (which is relevant for performing tasks) when dealing with imperfect models, as in the case of our complex humanoid robot.
Therefore, we introduce a virtual noise term to compensate for the mapping between the image and cartesian space. 
This correction reduces the cartesian error at the end effectors by 20\,\%.
\item We validate the calibration results on the real robot.
For this evaluation, we use an external tracking system. The final cartesian error at the end effectors is 3.9\,mm on average and 9.2\,mm in the worst case.
\item The procedure of collecting measurements with the robot's internal RGB camera and performing the calibration takes under 30 minutes. 
Required for the speed is a method to select the poses accounting for a clear view of the markers while allowing for a wide variety of joint configurations. 
\item The dataset, as well as a Python package to calibrate a general elastic robot, are provided\footnote{ 
You can find the dataset, videos of the measurements, additional details to the methods, and the code to calibrate an arbitrary elastic robot at\\ \href{\gitlink}{\gitlink}.}. 
The tool allows a combination of non-geometric forward kinematics with different custom measurement functions.
\end{itemize}

\section{Related Work}\label{sec:Related-Work}

\begin{figure}[tb]
	\centering
	\includegraphics[width=\linewidth]{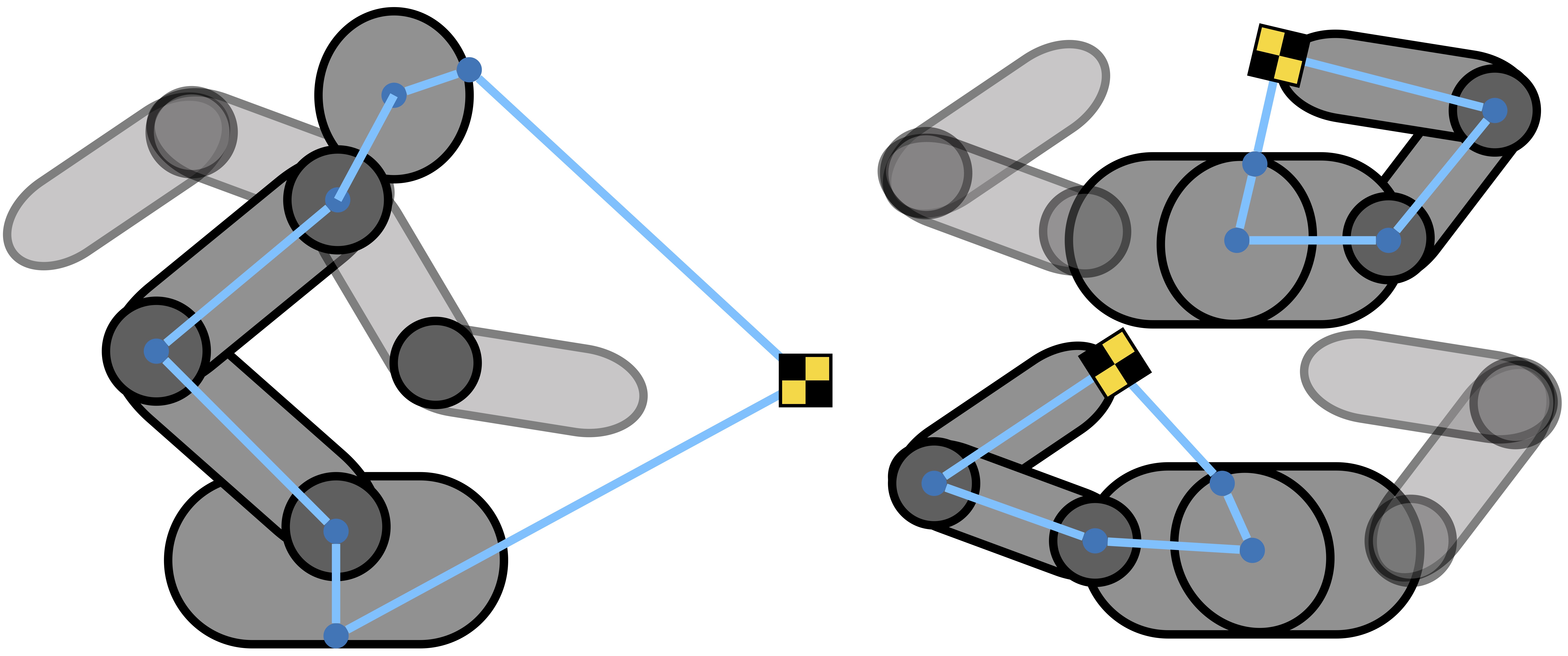}
	\caption{
Sketch of the calibration setup.
The robot collects images of markers on both of its hands and a pole in front of it. 
The blue chains show how forward kinematics plus camera projection close the measurement loop. 
Even if the arms are not directly involved in the pole measurements, their mass distribution in different positions influences the torso elasticities.}
	\label{fig:calibration_sketch}
\end{figure}

An accurate forward kinematics is relevant for most robotic applications; therefore, there are a lot of examples of successful calibrations.
Most of the time, the robotic arm's geometric model is calibrated with an external tracking system ~\cite{Ginani2011, Park2011, Xiong2017, VanWyk2019}. 
For an overview of the calibration and compensation of elastic robots, we refer to the related work in the preceding paper~\cite{Tenhumberg2021}. 

However, the calibration model must not only be expressive enough to match the real robot well. 
Another important aspect is a fast and easy calibration process to make it broadly applicable and easy to repeat if necessary.
Ideally, the robot uses its internal sensors to calibrate itself. 
This choice also ensures that precisely the same chain is calibrated the robot uses to perform its task.
\citet{Sang1996} introduced a self-calibration technique for active vision systems.
\citet{Hubert2012} added a bayesian approach and performed hand-eye calibration of an anthropomorphic robot using a checkerboard marker.

\citet{Maier2015} calibrated the joint offsets for the humanoid robot Nao
by following four checkerboard markers on both of its hands and feet with its RGB camera. 
Finally, \citet{Stepanova2019} used a combination of visual and tactile self-observing to calibrate all the DH parameters for the iCub robot. However, they did this only in simulation. 

While using a single camera for calibration is convenient, minimizing the error in image space is not the same as minimizing the relevant error in the cartesian task space.
Reprojection terms have been used for calibration with RGB-D cameras~\cite{Pradeep2014,  Ferguson2015}.
But they did not introduce virtual noise to account for an imperfect robot model and did not use it to transform the error in the image space to the task-relevant cartesian space. 
They only assumed a real noise for the joints, which can often be measured quite accurately and does not need to be handled as unknown noise.

We have already tackled the problem calibration for the humanoid Agile Justin~\cite{Bauml2014}.
Because this complex mechatronic system is built from lightweight components, it is especially susceptible to torque-dependent elasticity effects. 
Furthermore, its autonomous motion has strict accuracy requirements for its sensors and its forward kinematics.
In previous work, the multi-sensorial head~\cite{Carrillo2013}, the IMUs in the head and base~\cite{Birbach2014}, and the eye-hand chain~\cite{Birbach2015} were calibrated. 
While using the internal sensors, they ignored the torso for calibration, even as this chain has a significant error.     

In the preceding paper, we described the non-geometrical model of the humanoid Agile Justin and showed how to calibrate it with an external camera system~\cite{Tenhumberg2021}. 
Furthermore, we introduced an efficient technique to compensate for the implicit model, which was crucial as we wanted to use it in an optimization-based path planner~\cite{Wagner2013}.

The main drawback of this previous work was the dependence on the external tracking system.
This dependence not only limits the general applicability of the approach. Using the external system, we do not calibrate the relevant chain between the robot camera and body, which is pertinent to perform tasks autonomously. 
Instead, this consecutive paper provides a fast and accurate auto-calibration for Agile Justin's entire kinematic chain.
Relying only on the measurements collected by its internal RGB camera, we calibrate exactly the chain needed for whole-body motion planning. 

\begin{figure}[tb]
	\centering
	\includegraphics[width=\linewidth]{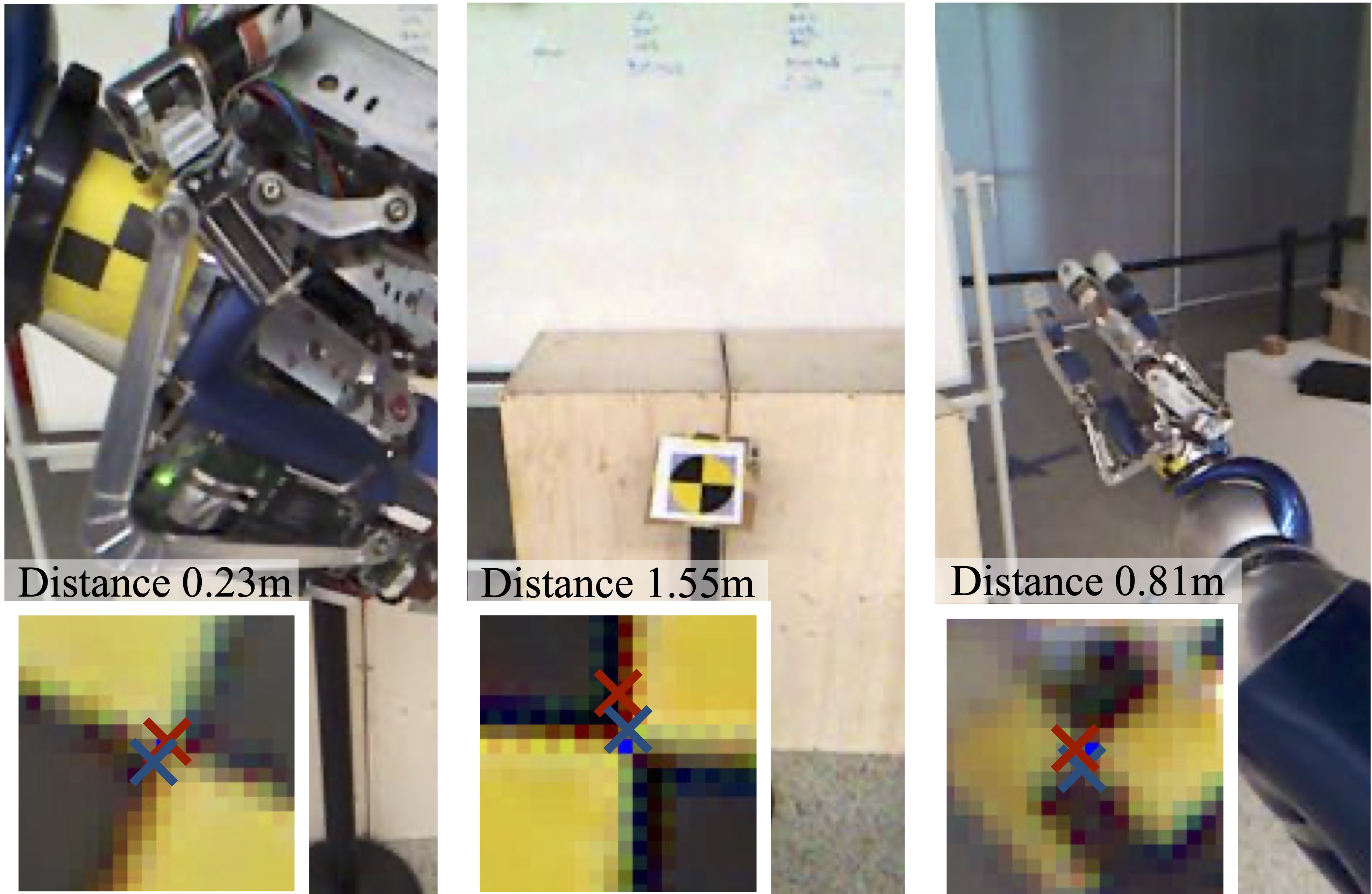}
	\caption{
DLRs's Agile Justin collects measurements to calibrate its non-geometric forward kinematics.
The images are from the robot's internal RGB camera with a resolution of $640\!\times\!480$, showing examples for the left arm, the right arm, and the pole.
The markers' distances to the camera vary between measurements from 0.2\,m up to 1.5m.
Without a correction (red), the pixel error is uniformly distributed over the distances, leading to more significant cartesian errors for detections further away from the camera as they correspond to a larger area.
The correction (blue) counteracts this and improves the cartesian accuracy by 20\,\%.}
	\label{fig:JustinCalibration}
\end{figure}

\section{Robot Model}\label{sec:Robot-Model}

\subsection{Forward Kinematics}

We use the same elastic forward kinematics as in~\cite{Tenhumberg2021} to model the robot.
To describe how the robots physical pose $\Trobot$ in the cartesian workspace changes with joint angles $\joints$ in the configuration space we us the DH formalism with the geometric parameter $\DHpar$.
The forward kinematics $f$ maps not only to the position of the end effector(s) but to all frames of the robot.
\begin{align}
    \Trobot = [\T{0}{1}, \T{0}{2}, \dotsc, \T{0}{N}] = f(\joints, \DHpar)
\end{align}
Following \citet{Caenen1990} we integrate the non-geometric effects from elasticities $\CPpar$ by explicitly expressing the influence of torques $\torque$ onto the DH parameters.
\begin{align}
\DHpar = \funDHpar(\DHparR, \CPpar, \torque) = \DHparR + \CPpar  \; \torque
\end{align}

The non-geometric forward kinematics is then given by
\begin{align*}
    \Trobot = \funForward(\joints, \CALparForward) = \funForward(\joints, \DHpar^*(\joints, \underbrace{\DHparR, \CPpar, \Masspar}_{\CALparForward})).
\end{align*}
where $\DHpar^*$ describes the solution of the non-geometric DH-parameters in torque equilibrium, resulting from the robots mass distribution $\Masspar$ in configuration q.
For more details on the derivation of those equations, as well as an algorithm to compensate and use this implicit model efficiently see \citet{Tenhumberg2021}.

\subsection{Camera Model}\label{sec:Camera-Model}

The forward kinematics describes the cartesian position of the different body parts.
If one wants to use a camera to measure those positions, one needs a model to project from the cartesian into the image space $\funCameraProjection\colon \mathbb{R}^3 \to \mathbb{R}^2$.
Here we use the classical pinhole model with radial distortion to project a 3D point of the marker $\markerPoint$ into 2D pixel coordinates $\markerProjection$~\cite{Birbach2015}.
First, the detected point $\markerPoint$ must be transformed into the camera frame $\Tcamera$. 
After this, the 3D point is projected along the z-axis of the camera frame with($\funCameraThreeTotTwo(\markerPoint)$
and radial distortion $\funCameraDistortion(\markerProjection, \CameraDistortion)$ is added. 
The pixel coordinates of the image $\markerProjection$ are then calculated as an offset from the cameras center point $\CameraCenterPoint$ scaled with the focal length $\CameraFocalLength$:
\begin{align}
   \funCameraProjection({}^{0}\markerPoint_{\mathrm{M}}, \Tcamera, \CALparProjection) = 
        \CameraCenterPoint + \CameraFocalLength \cdot \funCameraDistortion( 
            \funCameraThreeTotTwo(
                \TcameraInv \,\, {}^0\markerPoint_{\mathrm{M}}), \CameraDistortion), \\
    \mathrm{with} \;\; \funCameraThreeTotTwo(\markerPoint) = \left( \frac{\markerPoint_\mathrm{x}}{\markerPoint_\mathrm{z}}, \frac{\markerPoint_\mathrm{y}}{\markerPoint_\mathrm{z}} \right)^\mathrm{T} ; \; \funCameraDistortion(\markerProjection, \CameraDistortion) = \frac{\markerProjection}{1 + \CameraDistortion |\markerProjection|^2}
\end{align}
Besides the position of marker ${}^0\markerPoint_{\mathrm{M}} = \T{0}{\mathrm{i}} {}^{\mathrm{i}}\markerPoint_{\mathrm{M}} $ and frame of the camera $\Tcamera = \T{0}{\mathrm{j}}  \T{\mathrm{j}}{\mathrm{C}}$ relative to the forward kinematics $\Trobot$, the intrinsic parameters of the camera are also part of the calibration. Those additional parameters are combined and denoted as $\CALparProjection= [{}^i\markerPoint_{\mathrm{M}}, \T{j}{\mathrm{C}}, \CameraCenterPoint, \CameraFocalLength, \CameraDistortion]$.

\section{Calibration Problem} \label{sec:Calibration-Problem}

The goal of calibration is to find the set of parameters $\CALpar = [\CALparForward, \CALparProjection]$ which best fit the kinematic model and the camera model defined in the last section.

\subsection{Maximum a Posteriori Estimation}

As usual, we formulate the calibration as a probabilistic estimation problem~\cite{Bishop2007}. 
In \cref{fig:VirtualNoise} on the left, the probabilistic model is depicted using the functions introduced in \cref{sec:Robot-Model} to connect the input joint angles $\joints$ with the pixel coordinates of the markers $\markerProjection$ depending on the parameters $\CALpar$. 
The goal is to find the maximum of the posterior distribution $p(\CALpar | \DataSet)$ given the measurements $\DataSet = \{(\markerProjection^{(n)}, \joints^{(n)})\}_N$. 
For simplicity of notation, in what follows, we do not discern between the three different markers but assume that $\DataSet$ includes all measurements. 

As \cref{fig:VirtualNoise} shows, the stochastic variable of a markers pixel coordinates $\markerProjection$ can be expressed as a function (the so-called measurement function $\funMeasurement$) of the two input nodes $\joints$ and $\CALpar$ 
\begin{align*}
\funMeasurement(\joints, \CALpar) = 
\funCameraProjection( 
    \funForward(\joints, \CALparForward)_i {}^i\markerPoint_\mathrm{M} \,,\, 
    \funForward(\joints, \CALparForward)_j \T{j}{\mathrm{C}} \,,\, 
    \CALparProjection) \\
\markerProjection = \funMeasurement(\joints, \CALpar) + \eta_{\markerProjection}, \quad \eta_{\markerProjection} \sim N(0, C_{\markerProjection}), \quad C_{\markerProjection} = \sigma_{\markerProjection}^2 I.
\end{align*}
When the data is collected with the camera, there is real measurement noise on the pixels. 
We model this as gaussian noise $\eta_{\markerProjection}$ with zero mean diagonal variance $\sigma_{\markerProjection}$.
The maximum a posteriori (MAP) problem assuming a diagonal Gaussian prior $p(\CALpar)$ then results in a non-linear least squares problem
\begin{align}
\label{equ:map}
    \min_\CALpar \sum_n \log p(\markerProjection^{(n)}|q^{(n)}, \CALpar) + \log p(\CALpar)= \nonumber\\
    =\min_\CALpar \sum_n (\Delta u^{(n)})^\mathrm{T} C_\mathrm{m}^{-1} \Delta u^{(n)} +
    \Delta \CALpar^\mathrm{T} C_\mathrm{p}^{-1}  \Delta \CALpar,\\
 \text{with } \Delta u^{(n)} = \markerProjection^{(n)} - \funMeasurement(\joints^{(n)}, \CALpar), \quad C_{\mathrm{m}} = C_u,\\
 \text{and } \Delta \CALpar = \CALpar - \CALpar_{\mathrm{p}}, \quad C_\mathrm{p}= \operatorname{diag}{\sigma_{\mathrm{p}}^2}.\nonumber
\end{align}
 
In \cref{sec:Evaluation} we report the results of solving this optimization problem.

\subsection{Virtual Cartesian Noise}\label{sec:Virtual-Noise}

\begin{figure}[tb]
	\centering
	\includegraphics[width=\linewidth]{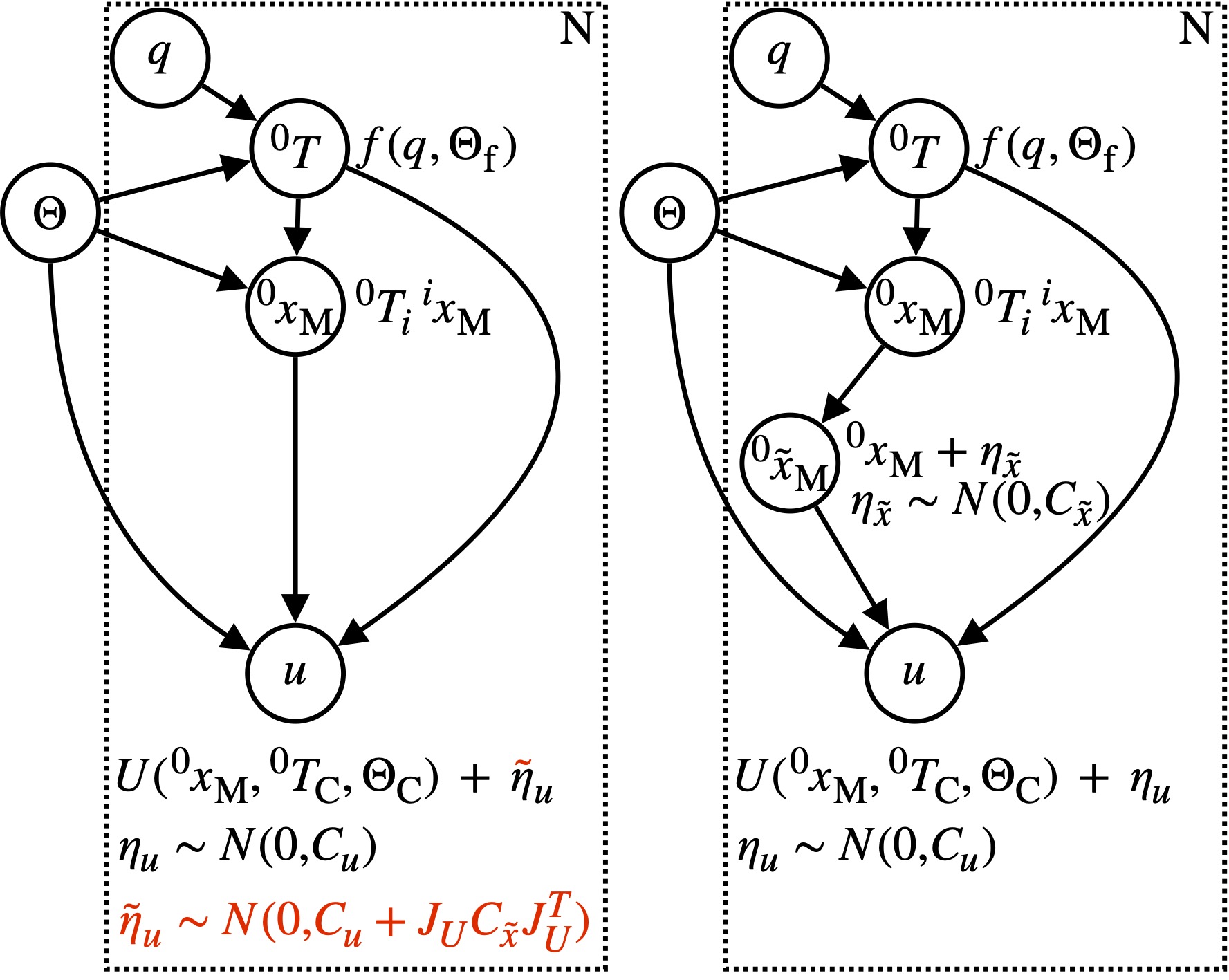}
	\caption{
    The probabilistic graph of the calibration problem includes the camera and robot model from \cref{sec:Robot-Model}. 
    It describes how the markers pixel coordinates $u$ are computed from the joint configuration $\joints$ and the model parameters $\CALpar$ for each of the $N$ samples. 
    \emph{Left (w/o red parts)}: In the original mapping, the real pixel measurement noise $\eta_u$ is the only source of stochasticity. 
    \emph{Right}: An additional virtual cartesian noise node is added to compensate for the imperfect (actually deterministic) kinematic model. 
    \emph{Left (with red parts)}: As shown in \cref{sec:Virtual-Noise}, the virtual noise can be incorporated into the original model, resulting in an effective pixel noise with a $\tilde{\sigma}_u$ depending on the distance of the marker to the camera ($\propto 1/z^2$).
}
	\label{fig:VirtualNoise}
\end{figure}

The MAP approach in \cref{equ:map} minimizes the difference between the measured marker and the reprojection of the marker in pixel coordinates. 
This method usually gives reasonable estimates for the parameters $\CALpar$ if the final pixel error gets very small, i.e., when the measurement model can fit the real robot well. 
But when fitting an imperfect model, i.e., a model which can not wholly reproduce all aspects of the real robot, the vanilla MAP approach still would minimize the overall pixel error by equally distributing the error between all measurements in \cref{equ:map}. 
Intuitively, this is not what we expect from a good fit: we want a fit that minimizes the error in cartesian space. 
A camera measures angles. 
This means a pixel further away corresponds to a larger area in physical space than a pixel closer to the camera. 
Therefore a pixel error for a marker far from the camera should count more than a pixel error close to the camera. 

To achieve this in a methodological sound way, we introduce an additional node in the graphical probabilistic model (\cref{fig:VirtualNoise}, right graph).
This addition explicitly models the (actually deterministic) imperfection as additional \emph{virtual noise} in the marker's 3D position. 
\begin{align*}
\rel{\tilde{x}}{0}{\mathrm{M}} = 
\rel{x}{0}{\mathrm{M}} + \eta_{\tilde{x}}, \quad \eta_{\tilde{x}} = 
N(0, C_{\tilde{x}}), \quad C_{\tilde{x}} = 
\sigma_{\tilde{x}}^2 I.
\end{align*}
It is important to note that the noise is added in base coordinates as we want the model error to be distributed equally in the world (and not, e.g., relative to some moving frame of the robot). 
The pixel coordinates of this noisy marker position now depend on this additional noise term
\begin{align*}
\markerProjection = U(\rel{\markerPoint}{0}{\mathrm{M}} + \eta_{\tilde{\markerPoint}}, \T{0}{\mathrm{C}}, \CALpar_\mathrm{C}) + \eta_u.
\end{align*}
By marginalizing over the new variable $\rel{\tilde{x}}{0}{\mathrm{M}}$, we get the effect of the additional virtual noise on the distribution of the pixel coordinates explicitly
\begin{align*}
p(\markerProjection|\rel{\markerPoint}{\mathrm{0}}{\mathrm{M}}, \T{0}{\mathrm{C}}, \CALpar) = 
\! \int\! p(\markerProjection|\rel{\tilde{\markerPoint}}{\mathrm{0}}{\mathrm{M}}, \T{0}{\mathrm{C}},\CALpar) 
p(\rel{\tilde{\markerPoint}}{\mathrm{0}}{\mathrm{M}}|\rel{\markerPoint}{\mathrm{0}}{\mathrm{M}})\mathrm{d} 
\rel{\tilde{\markerPoint}}{\mathrm{0}}{\mathrm{M}}.
\end{align*}
Due to non-linearities in $U$, the resulting distribution is non-Gaussian. 
However, we can approximate it with a Gaussian distribution by linearizing $U$ for the noise and using Gaussian arithmetic.
This results in almost the same form as before except for a new effective covariance $\tilde{C}_\markerProjection = \tilde{C}_u(\rel{x}{\mathrm{0}}{\mathrm{M}}, \T{0}{\mathrm{C}}, \CALpar)$ which now also depends on the marker's coordinates.
\begin{align*}
\markerProjection \approx U(\rel{x}{0}{\mathrm{M}}, \T{0}{\mathrm{C}}, \CALpar_\mathrm{C}) + \tilde{\eta}_{\markerProjection}, \quad
\tilde{\eta}_{\markerProjection} \sim N(0, \tilde{C}_u), \\
\tilde{C}_u(\rel{x}{\mathrm{0}}{\mathrm{M}}, \T{0}{\mathrm{C}}, \CALpar) = C_{\markerProjection} + J_U C_{\tilde{x}} J_U^{\mathrm{T}},\\
J_U(\rel{x}{\mathrm{0}}{\mathrm{M}}, \T{0}{\mathrm{C}}, \CALpar) = \frac{\partial U(x,\T{0}{\mathrm{C}}, \CALpar_\mathrm{C}) )}{\partial x}\Big|_{x=\rel{x}{0}{\mathrm{M}}} \end{align*}
Assuming that the camera distortion can be neglected for calculating the effective noise distribution, we finally get
\begin{align}
\label{equ:covariance}
\tilde{C}_{\markerProjection} \approx \sigma_u^2 I + \sigma_{\tilde{x}}^2 \left(\frac{f_{\mathrm{C}}}{x_\mathrm{z}}\right)^2 
\begin{pmatrix}
1 + \frac{x_\mathrm{x}^2}{x_\mathrm{z}^2} & \frac{x_\mathrm{x} x_\mathrm{y}}{x_\mathrm{z}^2} \\
\frac{x_\mathrm{x} x_\mathrm{y}}{x_\mathrm{z}^2} & 1 + \frac{x_\mathrm{y}^2}{x_\mathrm{z}^2} 
\end{pmatrix},
\end{align}
where $x = \rel{x}{\mathrm{C}}{\mathrm{M}} = \T{0}{\mathrm{C}}^{-1} (\rel{x}{\mathrm{0}}{\mathrm{M}})$.

The resulting MAP problem looks exactly the same as the original one \cref{equ:map}, except that the weighting of the individual measurement errors is changed to $C_\mathrm{m} = \tilde{C}_u(\rel{x}{\mathrm{0}}{\mathrm{M}}, \T{0}{\mathrm{C}}, \CALpar)$. 
Looking at \cref{equ:covariance}, this result means that markers further away are more critical and scaled with $z^{2}$ -- just as we intuitively expected.

\section{Efficient Sample Collection}\label{sec:Configuration-Selection}

\begin{figure}[tb]
	\centering
	\includegraphics[width=\linewidth]{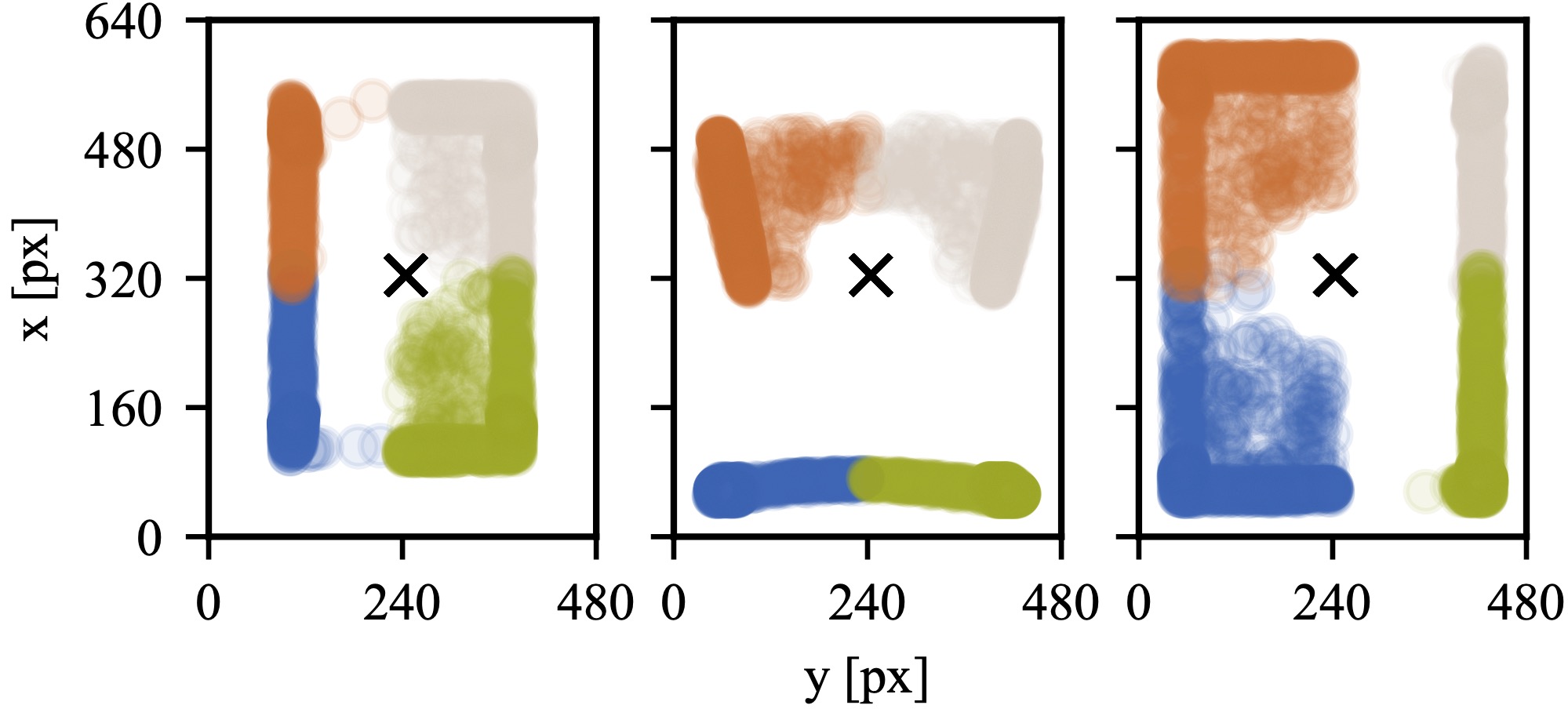} 
	\caption{
The different marker positions in the image for the left arm, the pole on the floor and the right arm. 
We move the pan-tilt joints of the robot's neck to get a good coverage of the image over all markers.
}
	\label{fig:markers_in_image}
\end{figure}

\begin{figure}[tb]
	\centering
	\includegraphics[width=\linewidth]{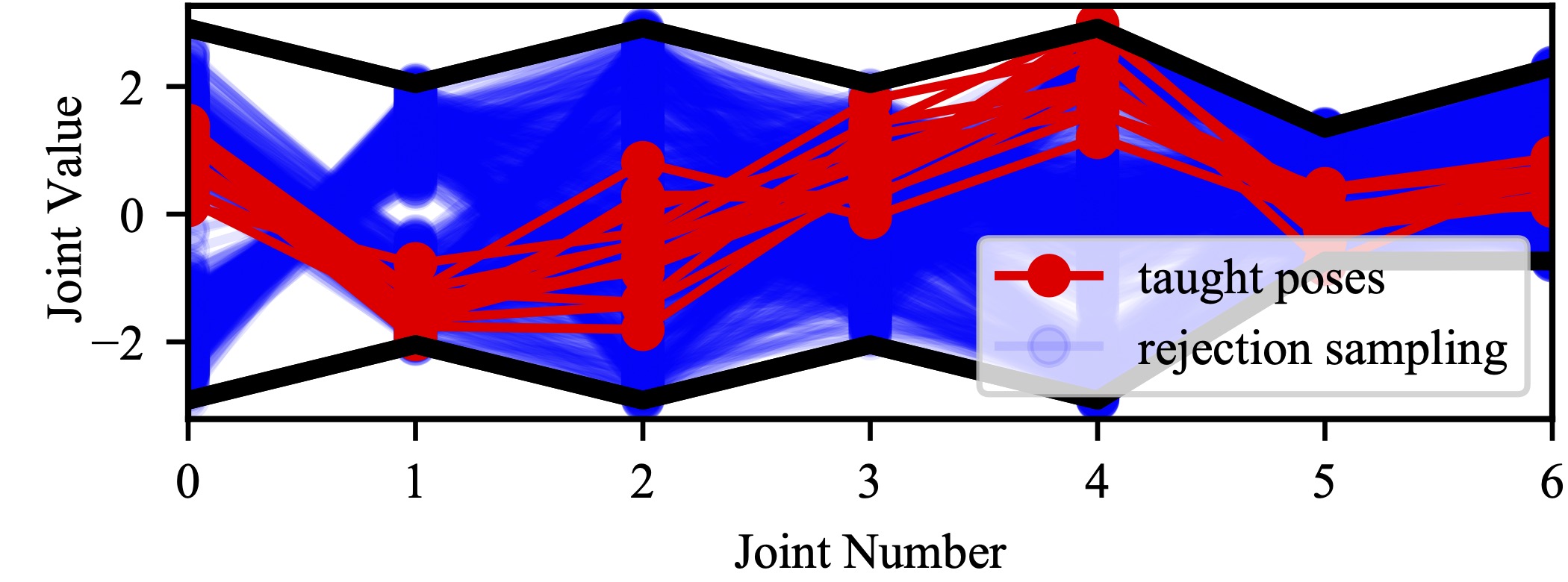}
	\caption{
The different distributions in configuration space of the right arm with 7 joints.
In red are the taught poses~\cite{Birbach2012, Carrillo2013, Birbach2015}. In blue are the configurations resulting from the rejection sampling approach described in \cref{sec:Configuration-Selection} leading to a broader distribution.
The joint limits are black.
}
	\label{fig:kinect-right_q_configurations}
\end{figure}

One goal of selecting measurement poses is to cover the whole configuration space. 
Sampling uniformly in the configuration space ensures that the calibrated model works well even if the robot moves autonomously and uses its full range of motion far away from taught standard configurations.

Nevertheless, the configurations must be feasible for the measurement setup. 
When the robot uses its camera to collect measurements of the markers, it imposes strict constraints. 
The markers must be in the camera's field of view, must not be occluded by the robot's own body, and must face toward the camera.
We check for occlusion with simple ray tracing and a sphere model of the robot by drawing a straight line from the camera to the marker and ensuring it does not collide with any of the spheres.
Checking if the camera and the marker face each other can be done by simply calculating the scalar product between their relative position and viewing direction. 
Furthermore, we want to ensure that only a single marker is visible to the robot at any configuration. 
In the case of Agile Justin, we needed over 10 million configurations to find 100 feasible measurement configurations for a marker.

As all those calculations happen before the calibration, one must account for uncertainties with larger thresholds and safety margins.
\cref{fig:kinect-right_q_configurations} shows that this general rejection sampling approach (blue) is better suited to get an even distribution over the configuration space. 
In contrast, in red are the taught poses used in prior works. 
This comparison shows that even for experts, it is hard to choose unbiased configurations for a complex humanoid.

We collect multiple measurements per robot configuration to speed up the calibration procedure.
In the case of Agile Justin, the camera is mounted on a pan-tilt joint, allowing us to adjust the marker's position in the image easily.
Assuming that in the initial configuration, the marker is roughly in the center of the camera's field of view, we move the camera to collect four additional measurements where the tag is in one of the corners of the image each time. 

\cref{fig:markers_in_image} shows the marker positions in the image when adjusting the head to collect multiple measurements per configuration. 
The main reason for only moving the head while keeping the rest of the body fixed is to increase the number of samples quickly.
Furthermore, good image coverage makes calibrating the camera intrinsics easier. 
To ensure that, we used neck joints to move the projection of the marker toward the corners while keeping a safety margin to the image center and borders. 

As in \citet{Tenhumberg2021}, we solve a traveling salesman problem 
to order the configurations we want to measure and reduce the time for calibration.
Furthermore, we use an optimization-based path planner to perform short and collision-free paths between the measurement configurations.

\section{Experimental Evaluation}\label{sec:Evaluation}

\begin{figure}[t]
	\centering
	\includegraphics[width=0.9\linewidth]{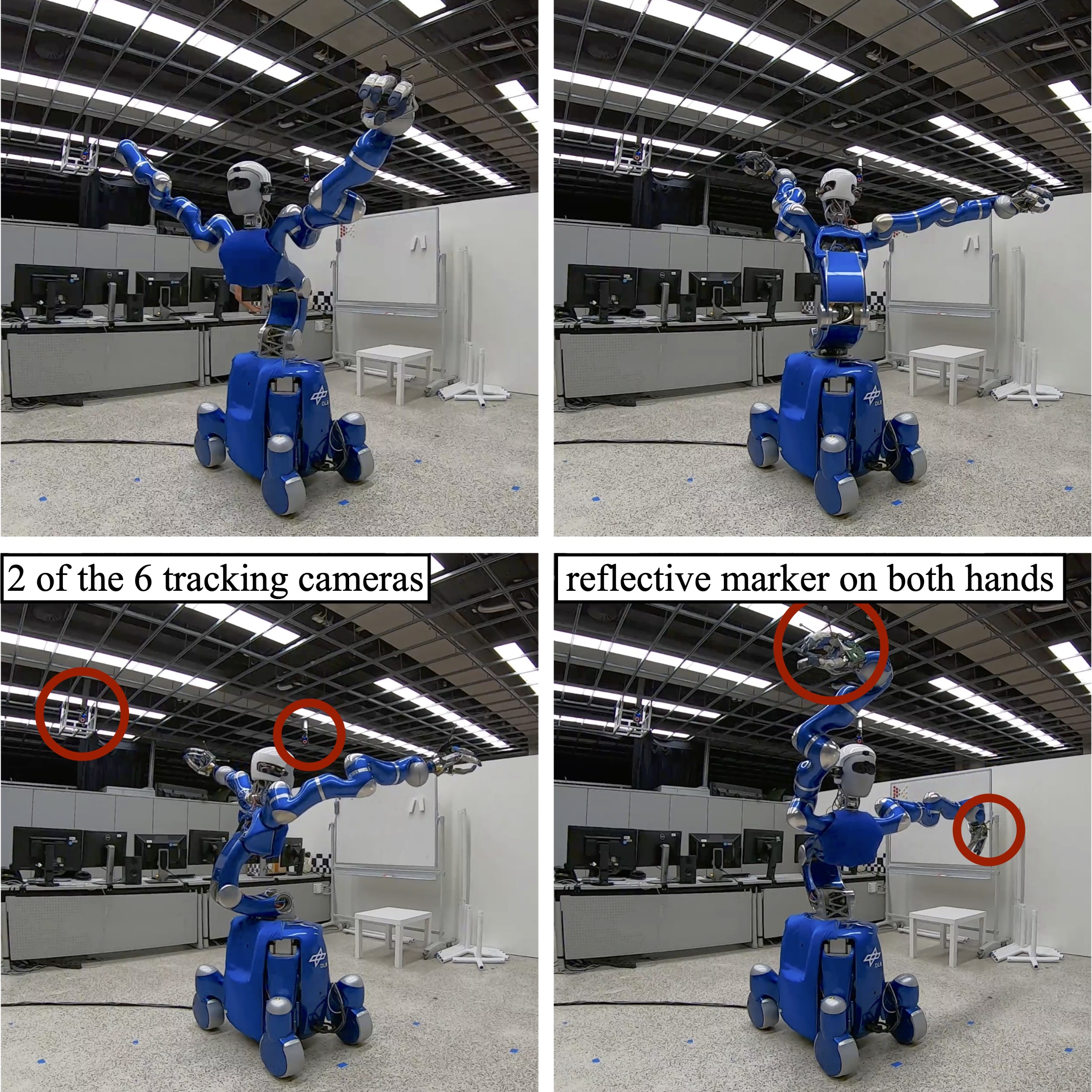}
	\caption{
We also collect measurements using the Vicon tracking system described in \cite{Tenhumberg2021} to evaluate our approach in the cartesian space.
This external tracking system consists of six cameras mounted on the ceiling and directly tracks the cartesian position of retro-reflective markers with high accuracy.
}   
    \vspace*{-2mm}
    \label{fig:ViconCalibration}
\end{figure}

\begin{figure}[t]
	\centering
	\includegraphics[width=\linewidth]{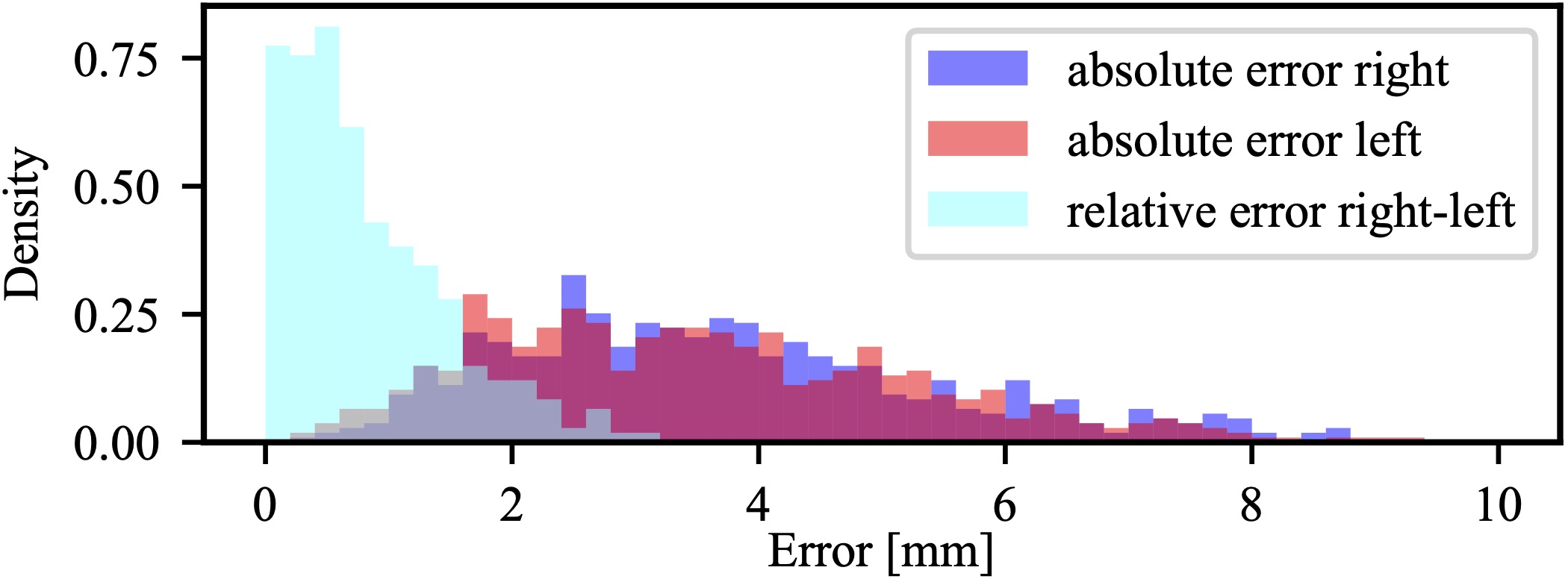}
	\caption{
Absolute cartesian error of the left and right arm after the full calibration. 
The relative error between those two is significantly smaller, indicating that the main part of the remaining error comes from the torso.
	}
	\vspace*{-2mm}
	\label{fig:difference_right_left}
\end{figure}

\begin{figure}[t]
	\centering
	\includegraphics[width=\linewidth]{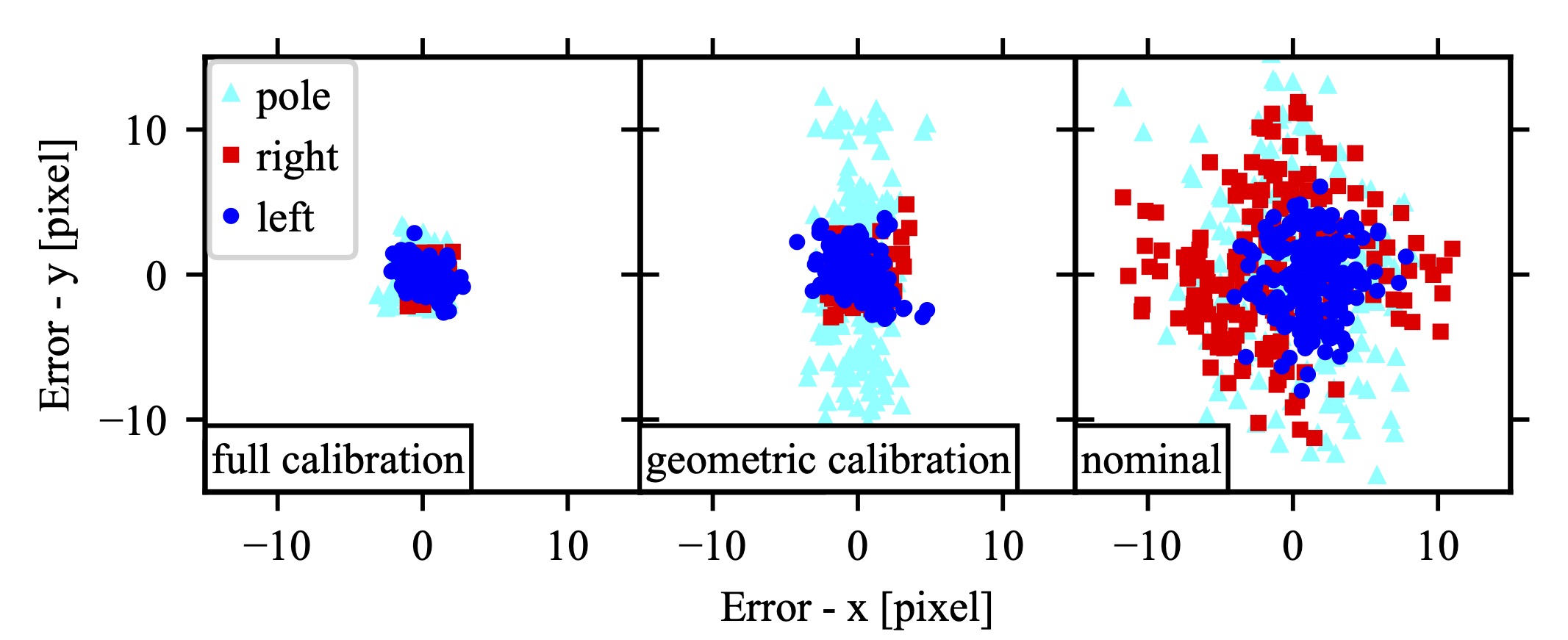}
	\caption{
Distribution of the errors in image space for the three markers at the pole, the left, and the right wrist. 
Calibrating only the geometric parameters (center) does not explain the elasticities in the torso seen in the pole marker.
The entire calibration (left) with geometric and non-geometric parameters distributes the remaining errors uniformly over the three markers.
	}
    \vspace*{-2mm}
	\label{fig:pixel_error_scatter}
\end{figure}

With this approach, we collected measurements for 50 configurations per marker, with five head positions each, resulting in $50 \times 5 \times 3 = 750$ samples for the three markers.
We split the set (equally for each marker) into 500 samples for calibration and 250 samples for evaluation. 
The virtual noise is set to $\sigma_{\tilde{x}} = 1\,$cm and the pixel noise to $\sigma_u=0.2$ (sub-pixel detection accuracy).
It takes roughly 8 minutes to collect the measurements for the markers on the hands and an additional 13 minutes to make the measurements for the pole.
The latter takes more time as the joint configurations are further apart in this setting, as the whole body is involved and not only one arm.
Together with performing the calibration itself (3 minutes), the whole procedure takes 32 minutes.
For comparison, the procedure with the external tracking system takes 25 minutes, which is a little faster as no separate configurations for each marker are necessary. 
However, there is an additional overhead for the setup of the tracking system.

For comparison, we also used the external cartesian tracking system to perform a new calibration using the method from \citet{Tenhumberg2021}. 
For this, we recorded recorded 100 different configurations (see \cref{fig:ViconCalibration}).
We also use this cartesian tracking data to evaluate the calibration based only on the head-mounted camera. 
For this, the position of the reflective targets and the tracking systems frame relative to the robot are recalibrated.

\cref{fig:pixel_error_scatter} shows the error in image space for the markers on the left and right arm and the pole for different calibration models. 
On the right-most image is the nominal forward kinematics, and in the center is the geometric model with the DH parameters. 
The left image shows the full calibration, including joint and lateral elasticities and the camera intrinsics.
One can see that the torso chain is responsible for significant non-geometric errors due to the large acting torques and its mechanical design with ropes.
In \cref{fig:difference_right_left} we further analyzed the influence of the different body parts.
The absolute cartesian errors of the right and left arm are red and blue, respectively. 
Here the torso chain is part of the measurements; therefore, its error is included.
However, the torso is excluded if we look at the distance between the left and right target and compare it against the measured length.
This reduced error in the relative arm positions indicates that the remaining error mainly comes from the torso chain.

The results also emphasize the need for our virtual noise term to cope with the imperfect model of the robot.
Even the elastic robot model does not capture all the relevant effects, and a significant error remains.
That the virtual noise helps to distribute the error evenly in the task-relevant cartesian space can be seen in \cref{tab:image_cartesian_error}. 
While the pixel error increases slightly, the mean and maximal cartesian error gets smaller by over 20\,\% when correcting for the mapping between the image and task space.
Furthermore, we show that it is possible to include the camera intrinsics $\CALparProjection$ in the calibration, further improving the accuracy.
The mean final error at the end effectors is 3.9\,mm on average and 9.2\,mm in the worst case. 
These results are comparable to a calibration using the cartesian measurements of an external tracking system, which is reported in the last row.

\begin{table}[t]
    \caption{
Error in the image and the cartesian space for different calibration models with and w/o virtual noise (VN), with and w/o intrinsic camera parameters ($\CALpar_\mathrm{C}$) and using only the camera (Image) or tracking system (Points).
}
    \vspace*{-2mm}
    \input{./tables/image_cartesian_error}
    \label{tab:image_cartesian_error}
\end{table}

\section{Conclusion}\label{sec:Conclusion}

The main advantage over the previous work from \citet{Tenhumberg2021} is that the new approach does not rely on an external camera system and calibrates the same chain used when performing manipulation tasks.
However, to achieve comparable accuracy in the cartesian task space, it is not enough to minimize the pixel error in the image of the single RGB camera.
We correct this mapping by introducing virtual cartesian noise. 
This way, it is ensured that the remaining error of our imperfect model is minimized in the task-relevant cartesian space and not the pixel space.
We show that this simple, self-contained approach leads to a similar good precision as using an external tracking system, reducing the error to 3.9\,mm on average.

In the future, we want to reduce the number of samples needed by optimizing the used kinematic configurations -- similarly to the work of \citet{Carrillo2013}, but for an elastic robot model and based on the here presented generic configuration generation scheme. 
We also want to improve the kinematic model by introducing an additional non-linear term for the torso chain, as we found here that this sub-chain is the primary source of the remaining error.

\footnotesize
\bibliographystyle{IEEEtranN-modified}
\bibliography{IEEEabrv, references.bib}
\end{document}

%% file: tables/image_cartesian_error.tex
\begin{tabular}{c|cc|ccc|ccc}
Calibrate    &            &        & \multicolumn{3}{c|}{Image Error {[}px{]}} & \multicolumn{3}{c}{Cartesian Error {[}mm{]}} \\ \cline{2-9} 
on             & $\Theta_C$ & VN  & $\mu$ & $\sigma$ & max                    & $\mu$        & $\sigma$        & max         \\ \hline
Images         & no         & no     & 1.05  & 0.59     & 3.76                   & 4.77         & 2.29            & 11.75       \\
Images         & yes        & no     & 0.97  & 0.53     & 3.44                   & 4.65         & 2.27            & 11.58       \\ \hline
Images         & no         & yes    & 1.21  & 0.70     & 4.13                   & 4.11         & 1.87            & 9.34        \\ 
Images         & yes        & yes    & 1.15  & 0.62     & 3.97                   & 3.94         & 1.83            & 9.16        \\ \hline
Points         & -          & -      & -     & -        & -                      & 3.12         & 1.71            & 8.23       
\end{tabular}

%% file: root.bbl
\begin{thebibliography}{19}
\providecommand{\natexlab}[1]{#1}
\providecommand{\url}[1]{#1}
\csname url@samestyle\endcsname
\providecommand{\newblock}{\relax}
\providecommand{\bibinfo}[2]{#2}
\providecommand{\BIBentrySTDinterwordspacing}{\spaceskip=0pt\relax}
\providecommand{\BIBentryALTinterwordstretchfactor}{4}
\providecommand{\BIBentryALTinterwordspacing}{\spaceskip=\fontdimen2\font plus
\BIBentryALTinterwordstretchfactor\fontdimen3\font minus
  \fontdimen4\font\relax}
\providecommand{\BIBforeignlanguage}[2]{{%
\expandafter\ifx\csname l@#1\endcsname\relax
\typeout{** WARNING: IEEEtranN.bst: No hyphenation pattern has been}%
\typeout{** loaded for the language `#1'. Using the pattern for}%
\typeout{** the default language instead.}%
\else
\language=\csname l@#1\endcsname
\fi
#2}}
\providecommand{\BIBdecl}{\relax}
\BIBdecl

\bibitem[Tenhumberg and B{\"{a}}uml(2021)]{Tenhumberg2021}
J.~Tenhumberg and B.~B{\"{a}}uml, ``{Calibration of an Elastic Humanoid Upper
  Body and Efficient Compensation for Motion Planning},'' in \emph{IEEE-RAS
  International Conference on Humanoid Robots}, vol. 2021-July, 2021.

\bibitem[B{\"{a}}uml et~al.(2014)B{\"{a}}uml, Hammer, Wagner, Birbach, Gumpert,
  Zhi, Hillenbrand, Beer, Friedl, and Butterfass]{Bauml2014}
B.~B{\"{a}}uml \emph{et~al.}, ``{Agile Justin: An upgraded member of DLR's
  family of lightweight and torque controlled humanoids},'' in
  \emph{Proceedings - IEEE International Conference on Robotics and
  Automation}.\hskip 1em plus 0.5em minus 0.4em\relax IEEE, 5 2014, pp.
  2562--2563.

\bibitem[Ginani and Motta(2011)]{Ginani2011}
L.~S. Ginani and J.~M. S.~T. Motta, ``{Theoretical and practical aspects of
  robot calibration with experimental verification},'' \emph{Journal of the
  Brazilian Society of Mechanical Sciences and Engineering}, vol.~33, no.~1,
  pp. 15--21, 3 2011.

\bibitem[Park and Kim(2011)]{Park2011}
I.~W. Park and J.~H. Kim, ``{Estimating entire geometric parameter errors of
  manipulator arm using laser module and stationary camera},'' \emph{IECON
  Proceedings (Industrial Electronics Conference)}, pp. 129--134, 2011.

\bibitem[Xiong et~al.(2017)Xiong, Ding, Zhu, and Su]{Xiong2017}
G.~Xiong, Y.~Ding, L.~M. Zhu, and C.~Y. Su, ``{A product-of-exponential-based
  robot calibration method with optimal measurement configurations},''
  \emph{International Journal of Advanced Robotic Systems}, vol.~14, no.~6, pp.
  1--12, 2017.

\bibitem[Van~Wyk et~al.(2019)Van~Wyk, Falco, and Cheok]{VanWyk2019}
K.~Van~Wyk, J.~Falco, and G.~Cheok, ``{Efficiently Improving and Quantifying
  Robot Accuracy In Situ},'' \emph{arXiv}, 8 2019.

\bibitem[{Sang De Ma}(1996)]{Sang1996}
{Sang De Ma}, ``{A self-calibration technique for active vision systems},''
  \emph{IEEE Transactions on Robotics and Automation}, vol.~12, no.~1, pp.
  114--120, 1996.

\bibitem[Hubert et~al.(2012)Hubert, Stuckler, and Behnke]{Hubert2012}
U.~Hubert, J.~Stuckler, and S.~Behnke, ``{Bayesian calibration of the hand-eye
  kinematics of an anthropomorphic robot},'' in \emph{2012 12th IEEE-RAS
  International Conference on Humanoid Robots (Humanoids 2012)}.\hskip 1em plus
  0.5em minus 0.4em\relax IEEE, 11 2012, pp. 618--624.

\bibitem[Maier et~al.(2015)Maier, Wrobel, and Bennewitz]{Maier2015}
D.~Maier, S.~Wrobel, and M.~Bennewitz, ``{Whole-body self-calibration via
  graph-optimization and automatic configuration selection},'' in \emph{2015
  IEEE International Conference on Robotics and Automation (ICRA)}, vol.
  2015-June, no. June.\hskip 1em plus 0.5em minus 0.4em\relax IEEE, 5 2015, pp.
  5662--5668.

\bibitem[Stepanova et~al.(2019)Stepanova, Pajdla, and Hoffmann]{Stepanova2019}
K.~Stepanova, T.~Pajdla, and M.~Hoffmann, ``{Robot Self-Calibration Using
  Multiple Kinematic Chains-A Simulation Study on the iCub Humanoid Robot},''
  \emph{IEEE Robotics and Automation Letters}, vol.~4, no.~2, pp. 1900--1907, 4
  2019.

\bibitem[Pradeep et~al.(2014)Pradeep, Konolige, and Berger]{Pradeep2014}
V.~Pradeep, K.~Konolige, and E.~Berger, ``{Calibrating a Multi-arm Multi-sensor
  Robot: A Bundle Adjustment Approach},'' in \emph{Springer Tracts in Advanced
  Robotics}, 2014, vol.~79, pp. 211--225.

\bibitem[Ferguson and Arora(2015)]{Ferguson2015}
M.~Ferguson and N.~Arora, ``{Robust and Efficient Calibration of Mobile
  Manipulators},'' Fetch Robotics, Tech. Rep., 2015.

\bibitem[Carrillo et~al.(2013)Carrillo, Birbach, Taubig, Bauml, Frese, and
  Castellanos]{Carrillo2013}
H.~Carrillo \emph{et~al.}, ``{On task-oriented criteria for configurations
  selection in robot calibration},'' in \emph{Proceedings - IEEE International
  Conference on Robotics and Automation}.\hskip 1em plus 0.5em minus
  0.4em\relax IEEE, 5 2013, pp. 3653--3659.

\bibitem[Birbach and Bauml(2014)]{Birbach2014}
O.~Birbach and B.~Bauml, ``{Calibrating a pair of inertial sensors at opposite
  ends of an imperfect kinematic chain},'' in \emph{2014 IEEE/RSJ International
  Conference on Intelligent Robots and Systems}, no. Iros.\hskip 1em plus 0.5em
  minus 0.4em\relax IEEE, 9 2014, pp. 422--428.

\bibitem[Birbach et~al.(2015)Birbach, Frese, and B{\"{a}}uml]{Birbach2015}
O.~Birbach, U.~Frese, and B.~B{\"{a}}uml, ``{Rapid calibration of a
  multi-sensorial humanoid's upper body: An automatic and self-contained
  approach},'' \emph{International Journal of Robotics Research}, vol.~34, no.
  4-5, pp. 420--436, 2015.

\bibitem[Wagner et~al.(2013)Wagner, Frese, and Bauml]{Wagner2013}
R.~Wagner, U.~Frese, and B.~Bauml, ``{3D modeling, distance and gradient
  computation for motion planning: A direct GPGPU approach},'' in \emph{2013
  IEEE International Conference on Robotics and Automation}, no. Iii.\hskip 1em
  plus 0.5em minus 0.4em\relax IEEE, 5 2013, pp. 3586--3592.

\bibitem[Caenen and Angue(1990)]{Caenen1990}
J.~Caenen and J.~Angue, ``{Identification of geometric and nongeometric
  parameters of robots},'' in \emph{Proceedings., IEEE International Conference
  on Robotics and Automation}.\hskip 1em plus 0.5em minus 0.4em\relax IEEE
  Comput. Soc. Press, 1990, pp. 1032--1037.

\bibitem[Bishop(2007)]{Bishop2007}
C.~Bishop, \emph{{Pattern Recognition and Machine Learning (Information Science
  and Statistics)}}.\hskip 1em plus 0.5em minus 0.4em\relax Springer, 2007.

\bibitem[Birbach et~al.(2012)Birbach, B{\"{a}}uml, Frese, Bauml, and
  Frese]{Birbach2012}
O.~Birbach \emph{et~al.}, ``{Automatic and self-contained calibration of a
  multi-sensorial humanoid's upper body},'' in \emph{2012 IEEE International
  Conference on Robotics and Automation}.\hskip 1em plus 0.5em minus
  0.4em\relax IEEE, 5 2012, pp. 3103--3108.

\end{thebibliography}
